\documentclass{article}
\usepackage{spconf}
\usepackage{booktabs}
\usepackage{amsmath,amssymb}
\usepackage{graphicx}
\usepackage[colorlinks=true,allcolors=blue]{hyperref}
\usepackage{xcolor}
\usepackage{microtype}
\usepackage{multirow}
\usepackage{placeins}
\usepackage{subfig}

\title{Reinforcing intra-operative tumour localisation with weak supervision from pre-operative appearances}

\name{%
\begin{tabular}{c}
    Sidhart Sami$^{1,3}$, Ali Sharjeel$^{1,3}$, Sri Atluri$^{1,2}$, Miao Zhao$^{1,2}$, \\
    Caroline Roney$^{1,2}$, Eirini Marouli$^{2,4}$, Shaheer U. Saeed$^{1,2,*}$
  \end{tabular}}%

\address{
$^{1}$Centre for Bioengineering, School of Engineering and Materials Science, \\Queen Mary University of London, London, United Kingdom\\
$^{2}$Digital Environment Research Institute, Queen Mary University of London, London, United Kingdom\\
$^{3}$National University of Computer and Emerging Sciences (FAST-NUCES), Karachi, Pakistan\\
$^{4}$William Harvey Research Institute, Faculty of Medicine and Dentistry,\\ Queen Mary University of London, London, United Kingdom\\
*Email: shaheer.saeed@qmul.ac.uk
}

\begin{document}
\maketitle

% ============================================================
% ============================================================
\begin{abstract}
Limited tumour visibility on intra-operative images complicates surgical navigation, including liver hepatocellular carcinoma (HCC) targeting on computed tomography (CT) imaging. This makes labelled datasets difficult to curate for deep learning, and transferring pre-operative delineations through registration remains challenging due to intra-operative deformation and lack of paired pre- and intra-operative data. We propose to transfer tumour appearances learned from pre-operative magnetic resonance (MR) images to guide localisation on intra-operative CT through weak supervision. A classifier is trained to classify tumour presence on pre-operative images. This is applied to windows on intra-operative images, and despite the domain gap, it assigns higher tumour-presence probabilities on average to tumour-containing windows. These weak cross-domain signals guide a reinforcement learning agent to localise tumours on intra-operative images. We compare direct transfer, few-shot adaptation and synthetic transfer (trained using unpaired MR and CT), which translates CT into synthetic MR before localisation. On an external HCC cohort, synthetic transfer outperformed other tested methods and CT-only baselines. Our framework enables cross-modality tumour localisation without paired MR-CT training images.\\
\end{abstract}

 
\section{Introduction}
 
Multi-modal imaging is central to modern clinical oncology, where complementary imaging modalities are acquired at different stages of care to maximise diagnostic accuracy and interventional precision \cite{boehm2022multimodal}. A prominent example is the combined use of pre-operative magnetic resonance (MR) imaging and intra-operative computed tomography (CT) imaging in oncology, where each modality serves a distinct clinical role~\cite{forner2018hepatocellular}. In liver oncology, pre-operative MR imaging provides soft-tissue contrast that supports tumour detection and delineation~\cite{chernyak2018liver}. Intra-operative CT is rapidly acquired for navigation, but tumour visibility can be limited \cite{mukhopadhyay2026segmenting, roberts2018imaging}. Lesions such as hepatocellular carcinoma (HCC) may therefore be difficult to distinguish from surrounding liver parenchyma \cite{mukhopadhyay2026segmenting}.
 
Transferring tumour localisation information from pre-operative to intra-operative imaging is therefore clinically important but technically challenging. Image registration offers a means of transferring pre-operative tumour delineations, but its application faces several challenges. Firstly, the limited availability of publicly accessible paired MR-CT images constrains the training of deep learning methods that require paired data~\cite{fu2020deepreg, cao2018registration}. Secondly, intra-operative liver displacement, respiratory motion and patient repositioning introduce non-rigid deformations that cannot be adequately captured by rigid alignment~\cite{fu2020deepreg}. Thirdly, some unpaired deep learning-based deformable registration approaches rely on corresponding anatomical labels to guide alignment \cite{fu2020deepreg}; obtaining reliable tumour labels in intra-operative CT is challenging for large datasets as lesion visibility is low, requiring expensive expert annotations~\cite{mukhopadhyay2026segmenting}. Directly learning HCC localisation on intra-operative CT is similarly constrained by the substantial expert effort required to curate labelled training data. These challenges limit both registration-based label transfer
and direct learning of localisation for lesions with low
CT visibility, such as HCC. For example, recent approaches achieve lesion Dice scores of only 0.16-0.19 when MR and CT images are registered using tumour labels \cite{mukhopadhyay2026segmenting}. 
These low tumour Dice scores contrast with reported anatomical structure Dice scores of 0.72, highlighting that broader anatomical alignment does not necessarily ensure accurate tumour label transfer \cite{mukhopadhyay2026segmenting, neuberger2020liverbiopsy}.

Rather than transferring patient-specific tumour coordinates through registration, we propose to transfer learned tumour appearance information through weak supervision. A tumour presence classifier is first trained on MR images with binary labels indicating HCC tumour presence or absence. This classifier is then applied on local windows of intra-operative CT images, where it is able to generate the tumour-presence classification probability (from logits). Our central hypothesis is that, despite the modality gap, the classifier retains sufficient discriminatory information for tumour-containing windows to receive higher tumour-presence classification probabilities on average than windows without tumour. These classification probabilities are a `weak' signal of cancer presence likelihood \cite{saeed2024competingpixels,tan2025spars, saeed2025reasoning}, with higher values indicating higher cancer presence likelihood within the window. Despite the domain gap to CT, the windows with cancer lead to higher probabilities on average compared to windows without cancer based on preliminary results from bootstrapped averages. The task of localising cancer in the CT image then becomes finding the areas where windows lead to higher classification probabilities on average. We formulate this localisation using reinforcement learning (RL), where an agent moves a window across the CT image, and receives a classification probability-derived reward signal guiding the search towards areas of higher cancer likelihood. Window-level classification probabilities are spatially aggregated using maximum probability retention to generate a tumour score map, thresholded to obtain a segmentation.

We investigate three strategies for deriving the weak supervision signal. Firstly, the MR-trained classifier is simply applied to the CT images, in a `zero-shot' setting. Secondly, the MR-trained classifier is fine-tuned on a few CT examples, in a `few-shot' setting. Lastly,  `synthetic transfer', uses a cycle generative adversarial network (GAN) to translate the CT image into synthetic MR before localisation, where the CycleGAN itself is trained using unpaired CT and MR images only. It is noteworthy that our proposed framework circumvents the need for paired data or tumour segmentation labels.

Our contributions are summarised: 1) a weakly supervised framework for transferring pre-operative tumour appearance information to intra-operative localisation, without requiring paired images or tumour masks in the zero-shot setting; 2) an RL-based tumour search strategy that uses the pre-operative image trained classifier's classification probabilities as weak signals to guide tumour localisation on intra-operative images; 3) comparative evaluation of zero-shot, few-shot and synthetic image transfer strategies for localising HCC in intra-operative CT using a pre-operative MR-trained tumour classifier.

% ============================================================

\section{Methods}

\begin{figure}
    \centering
    \includegraphics[width=0.85\linewidth]{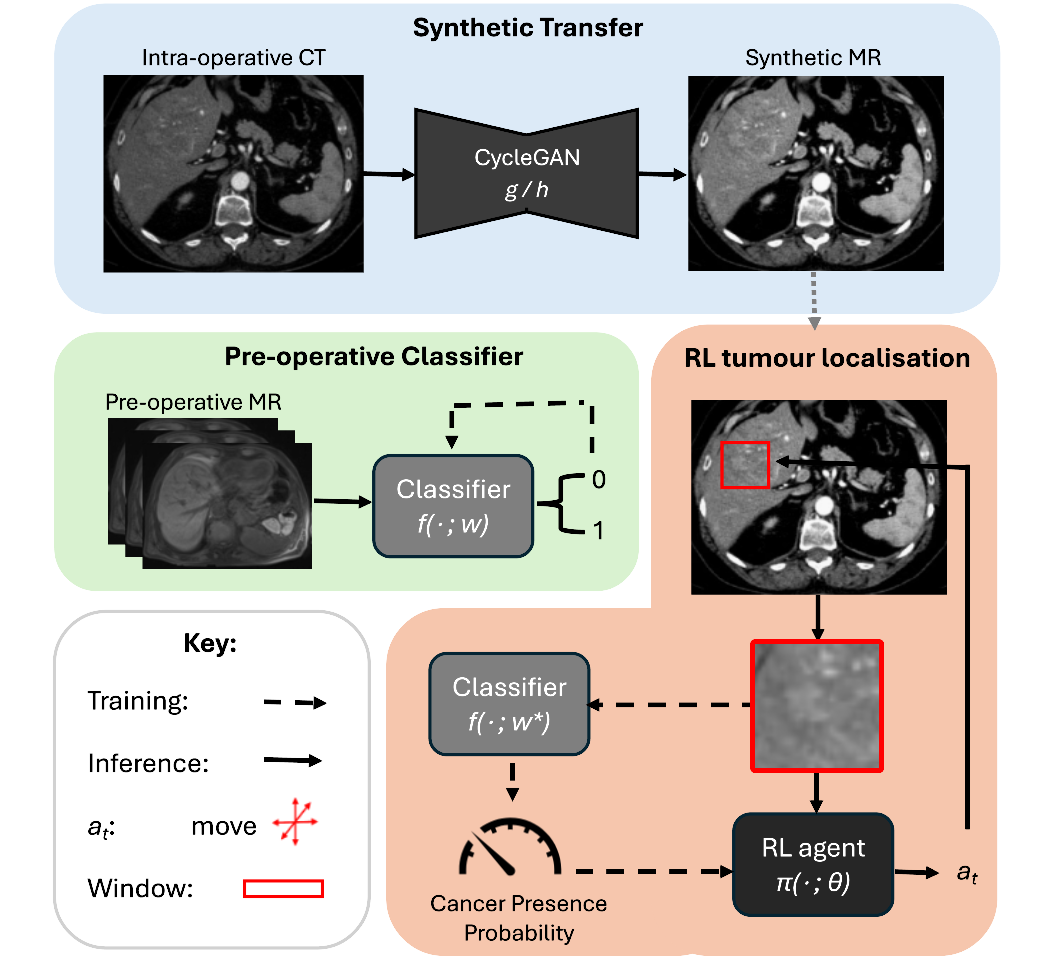}
    \caption{Summary of the proposed method.}
    \label{fig:method}
\end{figure}

\subsection{Tumour-presence classifier training}

We first train a classifier to recognise tumour presence in pre-operative MR images. Let $x \in \mathcal{X} = \mathbb{R}^{a \times b \times c \times d}$ denote an MR image input, where $a$, $b$ and $c$ are its spatial dimensions and $d$ is the number of channels, and let $y \in \mathcal{Y} = [0,1]$ denote its tumour-presence label.
Training labels are binary, with $y=1$ indicating tumour presence and $y=0$ indicating tumour absence. We define the classifier as:
\begin{equation}
    f(\cdot;w): \mathcal{X} \rightarrow \mathcal{Y},
\end{equation}
where $w$ denotes its trainable parameters and $f(x;w)$ is the predicted tumour-presence probability. The classifier is trained using binary cross-entropy loss $L(x,y; w)=-y\log f(x;w) -(1-y)\log\left(1-f(x;w)\right)$, with optimisation following:
\begin{equation}
        w^{*} = \arg\min_w
    \mathbb{E}_{(x,y)}
    \left[ L(x,y; w)
    \right],
\end{equation}
where the expectation is over labelled MR training inputs.

This objective requires only tumour-presence labels rather than voxel-wise tumour delineations.

\subsection{Window-based weak supervision}

To apply the same trained classifier $f(\cdot; w^*)$ to intra-operative CT, we extract local windows from each image, similar to \cite{saeed2024competingpixels, pocius2024weakly}. Let $z$ denote the full CT image and $z_p$ denote a fixed-size window centred at spatial position $p$ within the image. Each window is processed to match the input dimensions and preprocessing requirements of the classifier. Its tumour-presence classification probability is given by $f(z_p;w^{*})$.

Although the modality gap may affect the absolute classification probabilities, we hypothesise that tumour-containing windows retain higher values on average than windows without tumour \cite{pocius2024weakly,saeed2024competingpixels}. We therefore use these classification probabilities as relative weak signals for localisation. Different approaches to transfer this weak signal to CT are evaluated, as outlined in Sec. \ref{sec:methods_transfer}.

Classification probabilities from the windows visited during localisation are aggregated into a voxel-wise tumour score map $m$ using maximum probability retention. The localisation process is outlined in Sec. \ref{sec:methods_localisation}. Each visited window contributes its classification probability to every voxel within its spatial extent. For each voxel $v$, the map retains the highest probability among all visited windows containing that voxel::
\begin{equation}
    m(v) =
    \max_{t:\,v \in z_{p_t}}
    f(z_{p_t};w^{*}),
    \label{eq:aggregation}
\end{equation}
where $t$ indexes visited windows and $v \in z_{p_t}$ indicates that voxel $v$ lies within the spatial extent of the window. The map $m$ has the same spatial dimensions as the CT image $z$, with $m(v) \in [0,1]$ defined only at voxels covered by at least one visited window. The resulting map is converted into a tumour segmentation by thresholding, with the threshold $\tau$ selected through a grid search (see Tab. \ref{tab:tau}).

\subsection{Reinforcement learning for tumour localisation}\label{sec:methods_localisation}

We formulate window navigation as a sequential decision-making problem. At step $t$, the agent occupies position $p_t$ and observes the current CT window:
\begin{equation}
    s_t = (z_{p_t})
\end{equation}
where $z_{p_t} \in \mathcal{X}$ is the classifier input and $p_t$ specifies the three-dimensional window centre.

The action $a_t$ translates the window using $p_{t+1}=p_t+a_t$ by a fixed step size $\delta$ along one of the three image axes:
\begin{equation}
    a_t \in
    \left\{
        (\pm\delta,0,0),
        (0,\pm\delta,0),
        (0,0,\pm\delta)
    \right\}
\end{equation}

The agent receives a scalar reward $r_t \in [-1,1]$ based on the change in tumour-presence classification
probability:
\begin{equation}
    r_t =
    f(z_{p_{t+1}};w^{*}) - f(z_{p_t};w^{*}).
\end{equation}
Positive rewards encourage movement towards windows with higher tumour-presence classification probabilities.
The classifier remains fixed during policy optimisation.

The navigation policy $\pi(a_t \mid s_t;\theta)$, parameterised by $\theta$, assigns a probability to each
valid action given the current state. Its parameters are trained to maximise the expected discounted return:
\begin{equation}
    \theta^{*} =
    \arg\max_{\theta}
    \mathbb{E}_{\pi}
    \left[
        \sum_{t=0}^{T-1} \gamma^t r_t
    \right],
\end{equation}
where $T$ is the episode length and $\gamma \in (0,1)$ is the discount factor. Episodes begin at random valid
locations and terminate after $T$ steps. At inference, classification probabilities from the visited windows
are aggregated using Equation~\ref{eq:aggregation}.

\subsection{Enhancing the classifier-derived weak signal}\label{sec:methods_transfer}

\subsubsection{Zero-shot transfer}

The MR-trained classifier $f(\cdot;w^{*})$ remains fixed, and no labelled CT examples are used for classifier adaptation. This is applied directly to CT windows to obtain tumour-presence classification probabilities.

\subsubsection{Few-shot adaptation}

The MR-trained classifier is fine-tuned using a small set of CT inputs with binary tumour-presence labels. Fine-tuning uses the same binary cross-entropy objective, initialised with $w^{*}$. The adapted classifier is then fixed and used to obtain classification probabilities for CT windows during RL-based localisation.

\subsubsection{Synthetic image transfer}

An image translation function $g:\mathcal{X}_{\mathrm{CT}}\rightarrow \mathcal{X}_{\mathrm{MR}}$ is learned using an unpaired CycleGAN, without corresponding patient pairs or tumour labels. Training jointly learns $g$ and a reverse mapping $h:\mathcal{X}_{\mathrm{MR}}\rightarrow \mathcal{X}_{\mathrm{CT}}$. Adversarial losses encourage translated images to resemble the target modality, while cycle consistency encourages reconstruction of the inputs:
\begin{equation}
    L_{\mathrm{cyc}} =
    \mathbb{E}_{z}\left[\|h(g(z))-z\|_1\right]
    +
    \mathbb{E}_{x}\left[\|g(h(x))-x\|_1\right].
\end{equation}
The training objective combines the adversarial losses with $\lambda L_{\mathrm{cyc}}$, where $\lambda$ controls
the cycle-consistency contribution.

\begin{figure}
    \centering
    \includegraphics[width=0.9\linewidth]{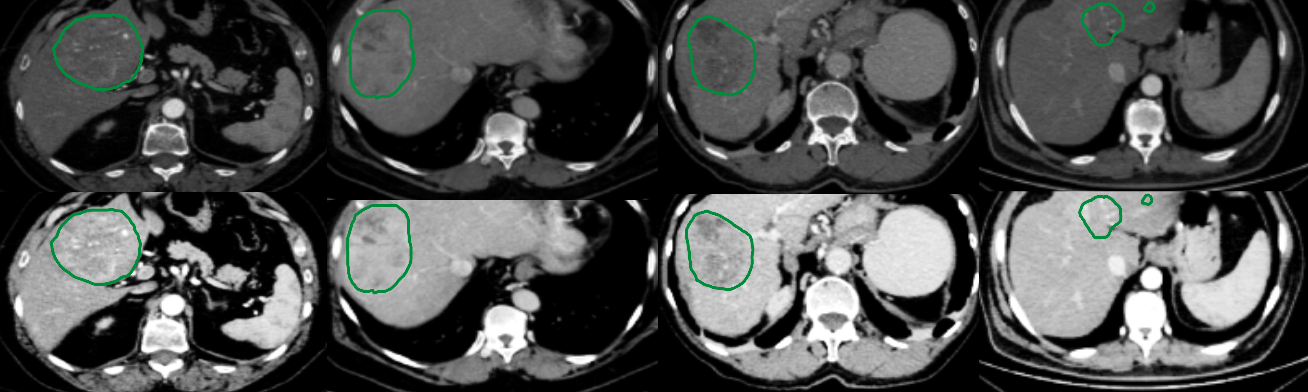}
    \caption{CycleGAN synthetic transfer (top: CT; bottom: MR) - average SSIM: 0.79 - ground truth segmentation in green}
    \label{fig:cyclegan}
\end{figure}

At inference, navigation is performed on $g(z)$, with the MR-trained classifier $f(\cdot; w)$ providing window-level classification probabilities for reward calculation and spatial aggregation. The resulting map is assigned to the corresponding CT coordinates (with anatomical correspondence retained owing to the spatial preservation enabled by the pixel-wise distance minimisation of the cycle-consistency loss \cite{zhu2017cyclegan} - as also noted in preliminary results presented in Fig. \ref{fig:cyclegan}).

% ============================================================
\section{Experiments}

\subsection{Data}

MR data: 334 patients from LLD-MMR~\cite{lldmmri}, ATLAS~\cite{atlas2023}, CHAOS~\cite{chaos2019}. All volumes were resampled to 1.0~mm isotropic spacing, cropped to 128$\times$128$\times$128 voxels, clipped at the 0.5th and 99.5th intensity percentiles, and z-score normalised.

CT data: 313 patients from WAW-TACE \cite{wawtace2024}, CHAOS-CT \cite{chaos2019}, and an internal dataset from [REC=14/LO/1264, IRAS=158321], including both pre- and intra-operative data. All volumes were windowed to -100 to +400 HU, resampled to 1.5~mm isotropic spacing, and z-score normalised. 

Both datasets were partitioned into development and validation sets using a 70\%/30\% split at the patient-level. Evaluation is reported on TCIA HCC-TACE-Seg (105 patients) \cite{hcctaceseg2023}, where volumes had radiologist voxel-wise tumour annotations as ground truth. We focus on HCC which is conventionally challenging to identify on intra-operative CT images \cite{mukhopadhyay2026segmenting}.

\subsection{Networks and hyperparameters}

The classifier used SE-ResNet3D \cite{hu2018senet} (learning rate $10^{-3}$, batch size 8). The CycleGAN used 3D encoder--decoder generators and multi-scale patch discriminators \cite{zhu2017cyclegan, hiasa2018crossmodality}. The RL agent used PPO \cite{schulman2017ppo} with a 3D convolutional network (learning rate $3\times10^{-4}$, discount factor 0.99). Hyperparameters provided in repository. On a single Nvidia Quadro GV100 GPU, the classifier converged in 8 hours, the CycleGAN in 16 hours, and the RL tumour localisation in 48 hours.

\subsection{Experimental setup}

We compared zero-shot, few-shot and synthetic transfer with recent methods~\cite{mukhopadhyay2026segmenting,fu2020deepreg}. Few-shot adaptation fine-tuned the classifier on 40 development CT volumes; synthetic transfer trained the CycleGAN on the MR and CT development sets before localisation on synthetic MR. Registration methods used paired evaluation data specified in~\cite{mukhopadhyay2026segmenting}, rather than our CT-only evaluation cohort. To ablate MR transfer, we trained CT-only RL baselines on the CT development set using rewards from tumour labels (full supervision; FS) or a CT-only classifier (weak supervision; WS). We included an FS nn-UNet trained on segmentation-labelled training data, as implemented in \cite{manghi2025deep}. Sensitivity analyses assessed window size and segmentation threshold $\tau$. We report patient-level Dice $\pm$ standard deviation, with t-tests for statistical comparisons. Localisation success rate is the proportion of individual discontiguous tumours across patients achieving Dice $>0.10$, following \cite{mukhopadhyay2026segmenting}.

\FloatBarrier

% ============================================================
\section{Results}
\label{sec:results}

\begin{figure}
    \centering
    \includegraphics[width=0.95\linewidth]{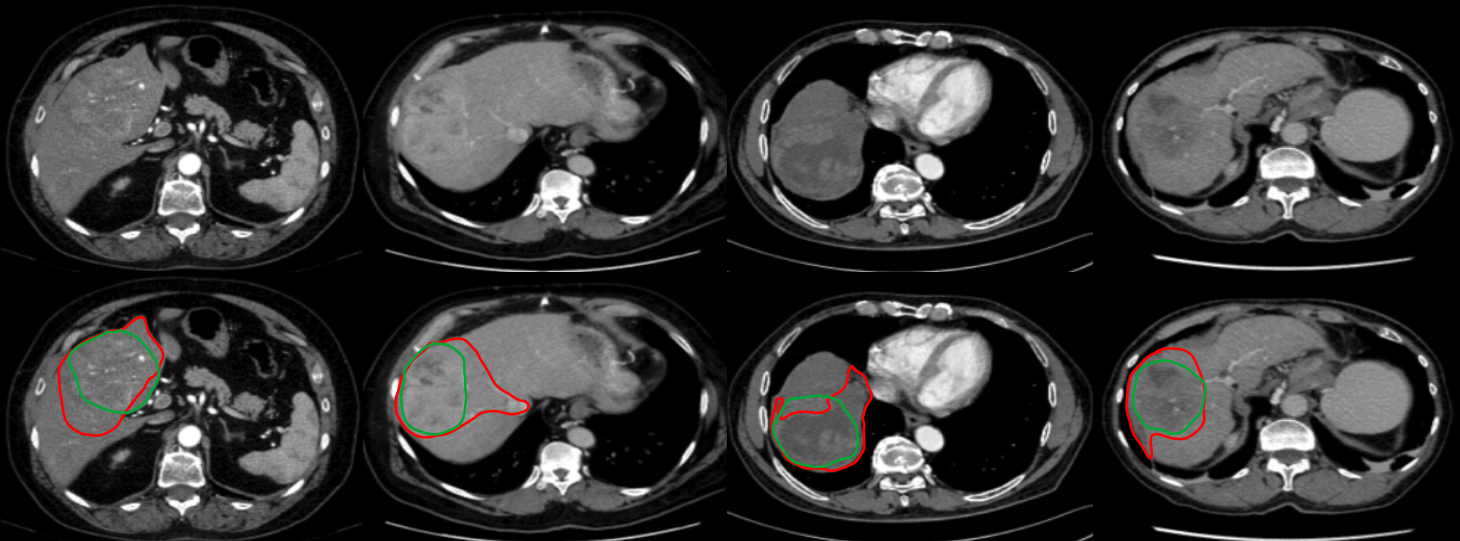}
    \caption{Qualitative results (top: images; bottom: ground truth segmentation in green and predicted in red - smoothed).}
    \label{fig:results}
\end{figure}

\subsection{Comparisons}

Table~\ref{tab:results} shows that synthetic transfer achieved the highest mean Dice score ($0.31\pm0.10$) and success rate (93\%), with significantly higher Dice than all other methods (all $p<0.01$). No statistically
significant differences in Dice were found between few-shot transfer and CT-only nn-UNet FS baseline ($p>0.11$).
Registration results are provided as contextual comparisons because they were obtained on a different evaluation cohort in the respective cited works. Qualitative results in Fig. \ref{fig:results}.

\begin{table}[!htbp]
\centering
\caption{Comparison with recent state-of-the-art.}
\label{tab:results}
\small
\begin{tabular}{@{}lcc@{}}
\toprule
Method & Mean $\pm$ SD & Success Rate \\
\midrule
zero-shot transfer & 0.16$\pm$0.06 & 0.67\\
few-shot transfer & 0.19$\pm$0.09 &  0.72\\
\textbf{synthetic transfer} & \textbf{0.31$\pm$0.10} & \textbf{0.93} \\
\hline
labelled deformable registration \cite{mukhopadhyay2026segmenting} & 0.19$\pm$0.06 & 0.71 \\
unlabelled deformable registration \cite{mukhopadhyay2026segmenting,fu2020deepreg} & 0.13$\pm$0.09 & 0.56\\
\hline
CT-only FS baseline & 0.07$\pm$0.05 & 0.49\\
CT-only WS baseline & 0.08$\pm$0.06 & 0.58 \\
CT-only nn-UNet FS baseline \cite{manghi2025deep} & 0.24$\pm$0.12 & 0.81\\
\bottomrule
\end{tabular}
\end{table}

\subsection{Sensitivity analysis}

We evaluated 16 candidate values per hyperparameter on the validation splits and report test results for the five best-performing settings. A $32\times32\times16$ voxel window achieved the highest mean Dice ($0.31\pm0.10$), closely followed by $32\times32\times32$ (Table~\ref{tab:window_size}). Mean Dice increased with threshold $\tau$ from 0.90 to 0.98, then decreased slightly at 0.99 (Table~\ref{tab:tau}).

\begin{table}[!ht]
    \centering
    \caption{Sensitivity analysis for hyperparameters.}
    \subfloat[Window size.\label{tab:window_size}]{%
        \begin{tabular}{c|c}
            \hline
            Window size & Mean $\pm$ SD \\
            \hline
            16$\times$16$\times$8 & 0.24$\pm$0.09\\
            \textbf{32$\times$32$\times$16} & \textbf{0.31$\pm$0.10}\\
            32$\times$32$\times$32 & 0.30$\pm$0.07\\
            48$\times$48$\times$24 & 0.28$\pm$0.11\\
            64$\times$64$\times$32 & 0.28$\pm$0.08\\
            \hline
        \end{tabular}%
    }%
    \hspace{0.9cm}
    \subfloat[Threshold $\tau$.\label{tab:tau}]{%
        \begin{tabular}{c|c}
           \hline
            $\tau$ & Mean $\pm$ SD \\
            \hline
            0.90 & 0.20$\pm$0.09\\
            0.92 & 0.26$\pm$0.11\\
            0.95 & 0.27$\pm$0.10\\
            \textbf{0.98} & \textbf{0.31$\pm$0.10}\\
            0.99 & 0.29$\pm$0.08\\
            \hline
        \end{tabular}%
    }%
\end{table}

\section{Conclusion}

Synthetic transfer significantly outperformed other evaluated methods, suggesting that reducing the modality gap strengthens MR-trained classifier guidance. Comparatively poorer CT-only results highlight the value of MR-derived information. Despite a 93\% localisation success rate, the Dice score of 0.31 indicates that HCC boundary delineation remains challenging and could be explored using alternative supervision. 

We developed a weakly supervised framework that uses an MR-trained classifier to guide liver tumour localisation on CT without paired MR--CT training images. This demonstrates the potential of transferring learned tumour appearances across modalities through weak supervision.

\section*{Open-source implementation}

The code has been open-sourced at:\\ \href{https://github.com/SidhartSami/cross-modality-weak-supervision}{github.com/SidhartSami/cross-modality-weak-supervision}.

\section*{Acknowledgements}
This work was supported by the International Alliance for Cancer Early Detection, an alliance between Cancer Research UK [EDDAPA-2024/100014], the University of Cambridge, the OHSU Knight Cancer Institute, the German Cancer Research Center (DKFZ), the University College London, the University of Manchester, and the Dana Farber Cancer Institute.

\section*{Compliance with Ethical Standards}

This research study was conducted retrospectively using human subject data made available in open access. The internal dataset from REC=14/LO/1264, IRAS=158321 complies with the Helsinki Declaration of 1975, as revised in 2000, and was obtained with informed patient consent.

\bibliographystyle{IEEEbib}
\bibliography{hcc_paper_draft}

\end{document}